\documentclass{article}
\usepackage{spconf,amsmath,graphicx}
\usepackage{amsmath,amssymb}
\usepackage[backend=biber,style=numeric]{biblatex}
\def\GK{{G_{\sigma}}}

\def\RR{{\mathbb{R}}}
\def\vs{{v_{\sigma}}}

\title{An Alternate View on Optimal Filtering in an RKHS}
\name{Benjamin Colburn$^{\star}$, Luis G. Sanchez Giraldo$^{\dagger}$,Jose C. Principe$^{\star}$\thanks{This material is based upon work supported by the Office of the Under Secretary of Defense for Research and Engineering under award number FA9550-21-1-0227, and partially supported by ONR grants N00014-21-1-2295 and N00014-21-1-2345.}}
\address{$^{\star}$University of Florida,$^{\dagger}$University of Kentucky}
\begin{document}
%
\maketitle
\begin{abstract}
Kernel Adaptive Filtering (KAF) are mathematically principled methods which search for a function in a Reproducing Kernel Hilbert Space. While they work well for tasks such as time series prediction and system identification they are plagued by a linear relationship between number of training samples and model size, hampering their use on the very large data sets common in today's data saturated world. Previous methods try to solve this issue by sparsification.  We describe a novel view of optimal filtering which may provide a route towards solutions in a RKHS which do not necessarily have this linear growth in model size. We do this by defining a RKHS in which the time structure of a stochastic process is still present. Using correntropy \cite{Santamaria2006}, an extension of the idea of a covariance function, we create a time based functional which describes some potentially nonlinear desired mapping function. This form of a solution may provide a fruitful line of research for creating more efficient representations of functionals in a RKHS, while theoretically providing computational complexity in the test set similar to Wiener solution.  
\end{abstract}
\begin{keywords}
RKHS, Nonlinear Filtering, Time-Series Prediction,Kernel Adaptive Filtering, Kernel Methods 
\end{keywords}
\section{Introduction}
\label{sec:intro}

Kernel Adaptive Filtering (KAF) methods extend known solutions for optimal linear filtering such as Least-Mean Squares (LMS) to the potentially infinite dimensional feature space emitted by a reproducing kernel. In the interest of brevity see \cite{manton2015} for a definition and explanation of reproducing kernels. For example, Kernel Least-Mean Squares (KLMS) extends the LMS solution to the feature space of a Reproducing Kernel Hilbert Space(RKHS). The optimal mapping function given by KAF algorithms is an inner product in a RKHS which is expressed as the linearly weighted sum of kernels centered at embedding vectors in $\RR^L$, 

\begin{equation}
   f(\textbf{x}_t) = \langle \Omega^*,\phi(\textbf{x}_t) \rangle_{K}  = \sum_{i=1}^{N}\alpha_i K(\textbf{x}_i,\textbf{x}_t).
    \label{KAFSol}
\end{equation}

\noindent Where N is the number of training samples, $\textbf{x}_i$ are training input samples, $\alpha_i \in \RR$ are parameters learned during the training of the model, and $\Omega^*$ denotes the optimal weight vector in the feature space of the RKHS. The obvious drawback of a solution with this form is that the model size grows linearly with the number of training samples. Many algorithms such as QKLMS \cite{Chen2012} and NICE \cite{Li2017} deal with this problem by sparsification of the functional i.e. they either condense the information from multiple training centers into one center (quantization), or split the sum into smaller sums only one of which needs to be evaluated for a given test point.

The solutions given by KLMS \cite{Liu2008} and KRLS \cite{Engel2004} resemble mapping functions from vectors in $\RR^L$, which represent windows in time, to $\RR$. They don't employ weight functionals based on time delays, as can be seen in (\ref{KAFSol}), which resemble a function with weight values based on locations in $\RR^L$. We would like to depart from this type of solution and instead search for solutions based on the time dynamics of a signal with weight values ascribed to time delays similar to an impulse response in linear filtering. 

Norbert Wiener’s 1949 work on minimum mean square estimation (MMSE) initiated the theory of optimum filtering \cite{Wiener1949}. The mathematics to solve integral equations, the Wiener-Hopf method \cite{Wiener1931}, were crucial to arrive at the optimal parameter function, however, the methodology is rather complex. In digital signal processing using finite impulse response filters, the Wiener solution coincides with least squares, as proven by the Wiener-Kinchin theorem \cite{Wiener1930}. Unfortunately, the solution still exists in the span of the input data i.e., the corresponding filter is linear in the parameters and therefore it is not a universal functional approximator. 

The Wiener solution gives the impulse response of the desired system. This impulse response fully specifies a FIR system. In this paper we will develop the idea of an impulse response in a RKHS. The goal of this line of thinking is to develop a different way to specify systems in a RKHS based on the time dynamics of a signal, and move away from graphical representations of functions which demand linear growth in model size. 

\section{Methods}
\subsection{The Random Gaussian RKHS}
In this section we introduce a RKHS which allows for the definition of time-based functionals. Its construction is based on the well known Gaussian kernel defined on a stochastic process domain, $\{X(t), t \in T\}$. We will call this RKHS $H_{RG}$. 




While the RKHS induced by the Gaussian kernel, or $H_G$, and $H_{RG}$ use the same positive definite kernel, there is an important distinction between the two. Elements in $H_{RG}$ are collections of Gaussian functions indexed by time. This means that the time structure of a random process is still present in elements of $H_{RG}$, allowing for the definition of time based functionals. In $H_G$ this is not the case because the domain is defined on arbitrary pairs of samples of the random process. Another way to think about this distinction is $H_{RG}$ is made of elements which represent small trajectories through $H_G$ defined by the largest delay of interest in the analysis.

\subsection{Correntropy as an Extension of Covariance}
Given a time series $\{X(t),t\in T\}$, the auto-correntropy function $v_{\sigma}(s,t)$ is defined as a function from $T \times T \rightarrow \mathbb{R}$ given by
\begin{equation}
\begin{split}
\vs(s,t)&=\mathbb{E}_{s,t}[G_{\sigma}(X(s),X(t))]\\ &=\int\int G_{\sigma}(X(s),X(t)) p_{s,t}(X(s),X(t))ds dt
\end{split}
    \label{Vst}
\end{equation}

where $\mathbb{E}_{s,t}[\cdot]$ denotes mathematical expectation over a pair of r.v. in the time series $\{X(t),t\in T\}$. While it is true that any Mercer Kernel, $\kappa(X(s),X(t))$, can be employed instead of the Gaussian kernel, the symmetry, scaling, and translation invariant properties of $G_{\sigma}(X(s),X(t))$, confer additional properties and interpretation to correntropy \cite{Liu2010}. If a stochastic process is strictly stationary, then auto-correntropy is only a function of lag($\tau$) \cite{Santamaria2006}. 

\begin{equation}
     v_{\sigma}(\tau)= \mathbb{E}_{t,t-\tau} [G_{\sigma}(X(t),X(t-\tau))]
     \label{Vtau}
 \end{equation}

Here we will introduce an analogy between $H_{RG}$, the span of a time series in $L^2$ ($L^2(\{X(t), t \in T\}))$, covariance function ($R(s,t)$), and correntropy function ($\vs(s,t)$). The auto-covariance function of a stochastic process can be defined as, 

\begin{equation}
R(s,t) = \mathbb{E}_{s,t}[X(s)X(t)].
\label{Rst}
\end{equation}

Similarly to correntropy if X(t) is wide sense stationary then the auto-covariance function is only a uni-variate function of lag. 

\begin{equation}
 R(\tau) = \mathbb{E}_{t,t-\tau}[X(t)X(t-\tau)]
 \label{Rtau}
\end{equation}

\noindent This uni-variate version of the auto-covariance function is a measure of the correlations between a stochastic process and its lagged versions, or the correlation between functions in $L^2(\{X(t),t \in T\})$.
Analogously the uni-variate version of auto-correntropy defined in (\ref{Vst}) is a measure of the correlations between a stochastic process and its lagged versions when projected into $H_{RG}$.

A very important feature of $H_{RG}$ is that it is congruent with correntropy. An element $U \in H_{RG}$ can be written as $U=\sum_{i}\alpha_i \GK(X(t_i),\cdot)$. The inner product between two elements , U and A, in $H_{RG}$ can be written as, 
\begin{equation}
\begin{split}
    \langle U,A\rangle_{H_{RG}}& =\mathbb{E}[\langle\sum_{i}\alpha_i G_{\sigma}(X(t_i),\cdot),\sum_{j}\beta_j G_{\sigma} (X(s_j),\cdot)\rangle]\\
    &=\sum_{i,j}\alpha_i \beta_j \mathbb{E}[G_{\sigma}(X(t_i),X(s_j))].
\end{split}
    \label{HRGinner}
\end{equation}

Where $t_i$ and $s_j$ are in the index set T. This means that there is a congruence between the inner product in $H_{RG}$ and autocorrentropy.

We can define the cross-covariance function between two stochastic processes, X(t) and Z(t), as  

\begin{equation}
P(\tau) = \mathbb{E}[Z(t)X(t-\tau)].
\label{Ptau}
\end{equation}

The cross-correntropy function can be defined as 

\begin{equation}
 P_v(\tau) = \mathbb{E}[\GK(Z(t),X(t-\tau))].
 \label{Pvtau}
\end{equation}

\subsection{Calculation of Weight Vectors}
In discrete time linear filtering we can find the optimal FIR description of a system by using the Wiener solution in the following way. Let $\textbf{R}$ be a $L \times L$ autocovariance matrix. Let $\textbf{P}$ be the L dimensional cross-covariance vector between and input signal, X(t), and desired response, Z(t). Then the optimal impulse response or weight vector $\textbf{W}^*$ is 

\begin{equation}
   \textbf{W}^*  = \textbf{R}^{-1} \textbf{P}.
   \label{RW*}
\end{equation}
This is the well known Wiener solution to the optimal linear filter and works in wide sense stationary settings. 

In \cite{Parzen1961} it is shown that the Wiener solution can be reinterpreted as an inner product in the RKHS defined by the auto-covariance function. This interpretation yields an equivalent solution to the Wiener solution, but in a RKHS. This gives some motivation for us to develop an analogous set of equations in $H_{RG}$ as optimal minimum mean square estimation methods in a RKHS are kernel agnostic. 

This analogous set of equation can be derived for stochastic processes in $H_{RG}$ in the following way. Let $\textbf{V}$ be the auto-correntropy matrix defined as, 

\begin{equation}
   \textbf{V} = \begin{bmatrix}
                \vs(0,0) & \hdots & \vs(0,L-1) \\
                \vdots & \ddots & \vdots \\
                \vs(L-1,0) & \hdots & \vs(L-1,L-1) \\
                \end{bmatrix}
    \label{Vmatrix}
\end{equation}

Let $\textbf{P}_v$ be the cross-correntropy vector defined as,

\begin{equation}
\textbf{P}_v = [P_v(0), P_v(1),\hdots, P_v(L-1)]^T.
\label{Pvec}
\end{equation}

Then the optimal weight vector can be found as, 

\begin{equation}
   \textbf{W}^{*}_{v} = \textbf{V}^{-1} \textbf{P}_v.
   \label{VW*}
\end{equation}

This is where the added complication of working in a functional space becomes apparent. Its not immediately obvious how to use this weight vector in $H_{RG}$. For example, suppose we are given a window in time, $\textbf{x}_t = [X(t),X(t-1),\hdots,X(t-L-1)]$. The optimal filter output is just the inner product between the weight vector and the input vector. Hence the optimal weight function should be the optimal weight function in $H_{RG}$ with the projection of the input vector $\textbf{x}_t$ in $H_{RG}$ or, 

\begin{equation}
   \hat{f}(t) =  \sum_{\tau=0}^{L-1} \textbf{W}^{*}_{v}(\tau) \GK(X(t-\tau),\cdot).
   \label{VW*app}
\end{equation}

The issue is that $\hat{f}$ is a functional in $H_{RG}$ we do not know where this functional should be evaluated. The functional can potentially take any value in some range defined by $\textbf{W}^{*}_{v}$. This suggests that the role $\textbf{W}^{*}_{v}$ is to define some range of values and does not fully specify an adaptive model in $H_{RG}$. In order to recover a scalar valued output we must choose where to evaluate this functional. 

\subsection{Recovering a Scalar}

In the previous section we introduce the issue of choosing a point to evaluate $\hat{f}$. Now we give a simple way to solve this problem. The goal of optimal filtering is to map the input signal, X(t), to a desired signal, Z(t). We will assume that the mapping function we are looking for depends only on L previous samples of X(t) and that X(t) is a strictly stationary stochastic process. Let S denote the set of all time windows of size L in the training set, that is $S=\{\textbf{x}_i, \forall i \in N\}$, where  $\textbf{x}_i = [X(i),\hdots,X(i-(L-1))]$. We need to find a $\textbf{p}_i$ such that 

\begin{equation}
    \hat{y}(t) = \sum_{\tau=0}^{L-1} \textbf{W}^{*}_{v}(\tau) \GK(\textbf{x}_i(\tau),\textbf{p}_i(\tau)) = Z(i).
    \label{ScalarFWF}
\end{equation}

We will refer to $\textbf{p}_i$ as the partner of $\textbf{x}_i$. We found one reliable way to estimate the value of this partner is to use the relationship between the desired signal, Z(i), and $\textbf{W}^{*}_{v}$. Suppose you want to find the partner for a given window of time, $\textbf{x}_i$. We can measure the distance between the desired output Z(i) and each weight in $\textbf{W}^{*}_{v}$ to get the vector
\begin{equation}
\textbf{g}_i=[\GK(\textbf{W}^{*}_{v}(0),Z(i)),\hdots,\GK(\textbf{W}^{*}_{v}(L),Z(i))].
\end{equation}
The Gaussian evaluation, $\GK(\hat{\textbf{p}}_i(\tau), \textbf{x}_i(\tau))$ must be proportional to to $\textbf{g}_i(\tau)$. Then the distance $|\hat{\textbf{p}}_i(\tau) - \textbf{x}_i(\tau)|$ must be inversely proportional to $\textbf{g}_i(\tau)$. Essentially if the desired output is closer to a given weight value then we want to weight that value more heavily. We can then calculate how far from the center of a Gaussian kernel we need to evaluate it in order for $\GK(\hat{\textbf{p}}_i,\textbf{x}_i(\tau)) = \textbf{g}_i(\tau)$. Then
\begin{equation}
\hat{\textbf{p}}_i(\tau) =\textbf{x}_{i}(\tau) - \alpha \GK^{-1}(\textbf{g}_i(\tau)),
\end{equation}
where $\alpha$ is introduced as a hyper-parameter and controls the scale of the prediction, and the $\GK^{-1}(\cdot)$ is the inverse of the Gaussian kernel. Note that the symmetry of the Gaussian kernel means there are two possible values of $\GK^{-1}(\cdot)$ for any given value of $\textbf{g}_i(\tau)$. We can just pick one of these values since the choice will make no difference in the prediction of the model. Let $S^*$ denote the set of all partners in the training set, $S^* = \{\hat{\textbf{p}}_i, i \in N\}$.

Once we have this set of partners, $S^*$, we can make a prediction in the test set in the following way. First, we use a nearest neighbor search between a current test window, $\textbf{x}_t$, and all training windows in S. Then we can use the partner $\hat{\textbf{p}}_i$ of the closest sample in the training set to estimate a scalar output for the given test point. The final prediction using this method is given as 

\begin{equation}
    \hat{y}(t) = \sum_{\tau=0}^{L-1}\textbf{W}^{*}_{v}(\tau)\GK(\hat{\textbf{p}}_i,\textbf{x}_t(\tau))
    \label{FWFNNPred}
\end{equation}

where $\hat{\textbf{p}}_i$ is the partner of the closest training set sample. We can also use a K-nearest type approach where we find the K nearest training samples and average the output of these filters. In this case the final prediction given by the filter is, 

\begin{equation}
    \hat{y}(t)= \frac{1}{K} \sum_{k=0}^{K}\sum_{\tau=0}^{L-1}\textbf{W}^{*}_{v}(\tau)\GK(\hat{\textbf{p}}_i(\tau),\textbf{x}_t(\tau)) 
    \label{Knearest}
\end{equation}

Experimentally, we have found that averaging the results from the K nearest neighbors is usually beneficial for test set performance provided the correct number for K is selected. Note that alpha changes the scale of the output of the filter and should be chosen such that training set error is minimized. Also the prediction given by this method may have a bias, since we usually assume that the desired signal has a mean of zero we can simply subtract the mean out of the prediction to correct for this bias. We will refer to this method as the functional Wiener filter with nearest neighbor approach or $FWF_{NN}$. It's important to note that in the training set the computational complexity of this method is on the same order as the linear Wiener filter. Even with added computational complexity of the search at test time the $FWF_{NN}$ is more computationally efficient than its traditional KAF counterparts. 


\begin{table}[h]
\begin{center}
\caption{Computational Complexity: i = iterations, N= number of training samples, L=filter order, K= number of neighbors}
\label{tab1}
\begin{tabular}{|c|c|c|} \hline
\textbf{Filter} & \textbf{Complexity} & \textbf{Memory}\\ \hline
KLMS & $O(i)$ & $O(i)$ \\ \hline
KRLS & $O(i^2)$ & $O(i^2)$ \\ \hline
$FWF_{NN}$ Train & $O(L^2N) + O(N)$ & $O(2N+L^2)$\\ \hline
$FWF_{NN}$ Test & $O(KL)+O(log(N))$ & $O(2NL+L^2)$\\ \hline
\end{tabular}
\end{center}
\end{table}

\section{Experiments}


\subsection{Prediction of Mackey-Glass Time Series}

\begin{figure}[h]
\centering
\includegraphics[width=0.45\textwidth]{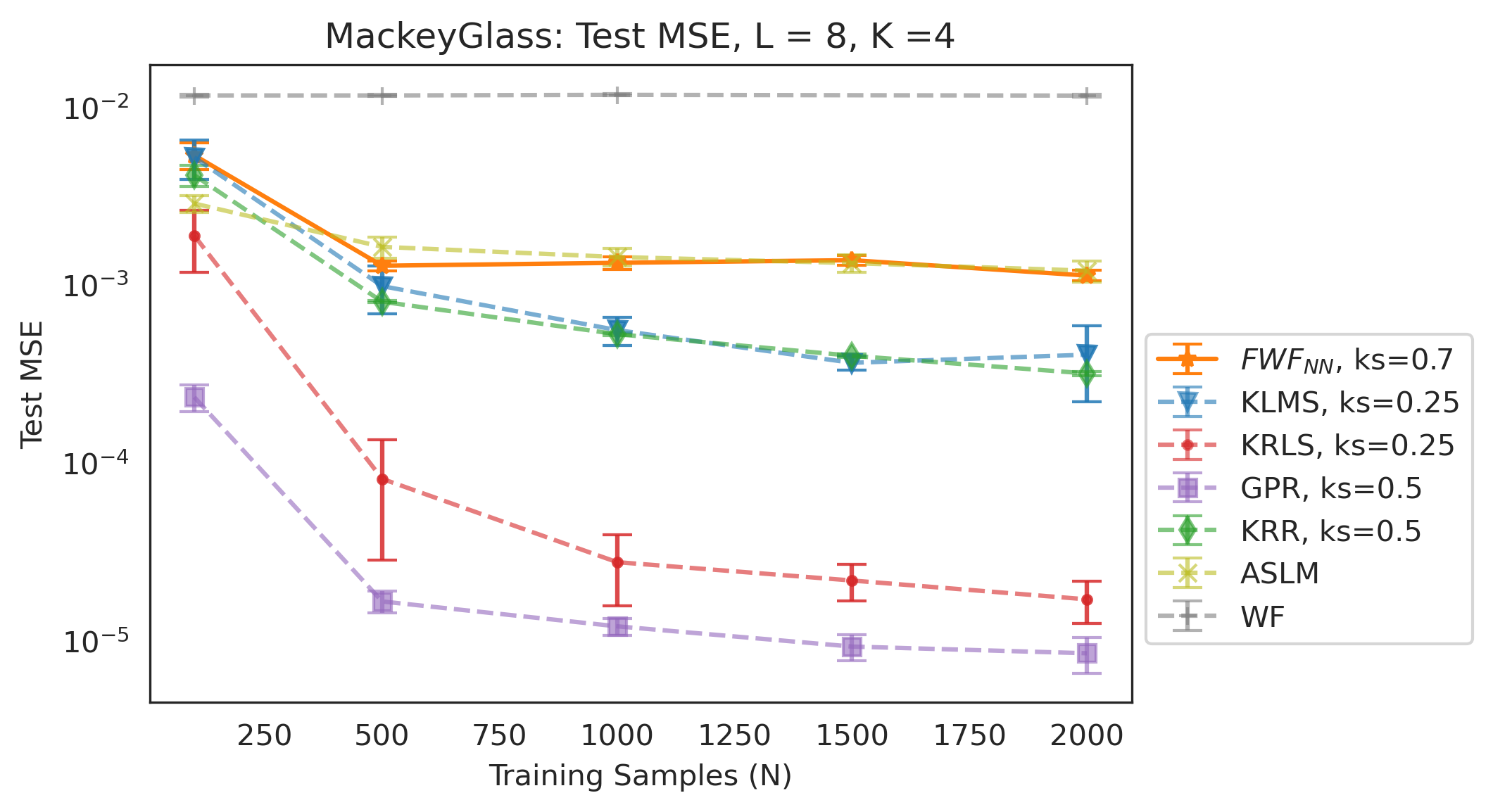}
\caption{Comparisons of test set MSE as a function of the number of samples in the training set (N). The error bars refer to a standard deviation of the test MSE.}
\label{MG_Comp}
\end{figure}
 
The Mackey-Glass (MG) times series is a nonlinear chaotic time series. In figure \ref{MG_Comp}, the performance of $FWF_{NN}$ is compared with other well known non-linear regression methods on the task of prediction of the Mackey-Glass time series. The methods we used for comparison were Gaussian Process Regression (GPR) \cite{Williams1995}, KRLS \cite{Engel2004}, KLMS \cite{Liu2008}, Kernel Ridge Regression (KRR) \cite{Hoerl2000}, Augmented Space Linear Regression (ASLM) \cite{Zhengda2020}, and the linear Wiener Filter (WF) \cite{Wiener1949}. The value of alpha used for the $FWF_{NN}$ was 0.445. The closest two neighbors were used to make predictions using equation \ref{Knearest}. Five-fold cross validation was used with 200 test samples to give some range of test set MSE achieved by each method. The $FWF_{NN}$ performed better than its linear counterpart, but is worse than the other nonlinear filtering methods. The average MSE at N=2000 in the training set for the $FWF_{NN}$ was $\approx1e-5$ this suggest that the model fails to generalize well. Finding a more robust approach for identifying partners during test time may improve performance.



\subsection{Prediction of Lorenz Time Series}
The Lorenz system is a well-known system introduced in \cite{Lorenz1963}. We use the $x$ component of the Lorenz attractor and to make the problem harder, the models predict $x(t+10)$ e.g. 10 samples ahead using the last $L$ samples. 
The value of alpha for the $FWF_{NN}$ was 0.355. Similar to with Mackey-Glass we will compare performance on the test set across 5-folds to other nonlinear regression models.

\begin{figure}[h]
\centering
\includegraphics[width=0.45\textwidth]{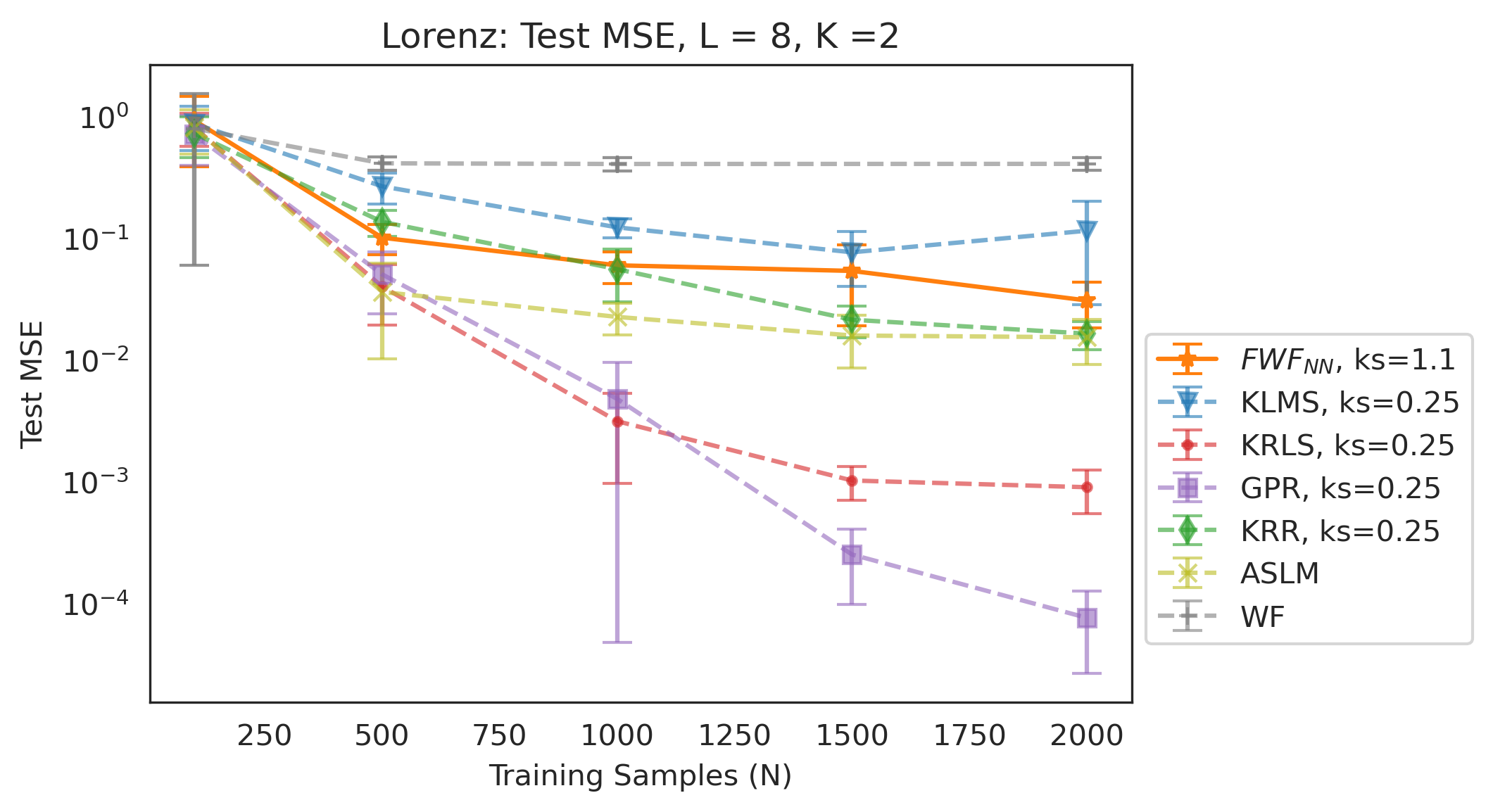}
\caption{Comparisons of test set MSE as a function of the number of samples in the training set(N). The error bars refer to a standard deviation of the test MSE.}
\label{LO_Comp}
\end{figure}


\section{Conclusions}

The main objective of this paper is to explore the idea of time-based functionals in a RKHS. We define the Random Gaussian RKHS which retains the time information in a stochastic process. We then use the correntropy function as an extension of the covariance function to find a weight function which contains the statistics of the auto-correntropy function of an input stochastic process in $H_{RG}$ and the cross-correntropy between the input stochastic process and the desired output. This is where we encounter an issue, the resulting weight vector does not fully specify a system in $H_{RG}$. We develop a heuristic method, $FWF_{NN}$, which gives a solution to this problem. 


The method still requires a search through training examples. While methods such as ball-tree can make this search efficient, computational complexity of the model evaluation at test time is still tied to the number of training examples.  The results are reasonable, but further research is needed to develop other methods for implementation which may improve generalization of the model and completely decouple evaluation computational complexity from number of training examples. Perhaps the more important contribution of this paper is to present a novel way to approach optimal nonlinear filtering in a RKHS. The idea of using time-based functionals in a RKHS to identify a desired system gives a new possible route to eliminate the linear growth in model size that plagues current KAF and kernel regression methods.   

\printbibliography

\end{document}